%% file: main.tex
\def\year{2022}\relax
\documentclass[letterpaper]{article} 
\usepackage{aaai22}  
\usepackage{times}  
\usepackage{helvet}  
\usepackage{courier}  
\usepackage[hyphens]{url}  
\usepackage{graphicx} 
\usepackage{natbib}  
\usepackage{caption} 
\DeclareCaptionStyle{ruled}{labelfont=normalfont,labelsep=colon,strut=off} 
\usepackage{algorithm}
\usepackage{algorithmic}

\usepackage{newfloat}
\usepackage{listings}
\floatstyle{ruled}
\newfloat{listing}{tb}{lst}{}
\floatname{listing}{Listing}
\usepackage{color} 

\title{
AI Explainability 360: Impact and Design
}
\author{
Vijay Arya, Rachel K. E. Bellamy, Pin-Yu Chen, Amit Dhurandhar, Michael Hind, Samuel C. Hoffman, Stephanie Houde, Q. Vera Liao, Ronny Luss, Aleksandra Mojsilovi\'c, Sami Mourad, Pablo Pedemonte, Ramya Raghavendra, John Richards, Prasanna Sattigeri, Karthikeyan Shanmugam, Moninder Singh, Kush R. Varshney, Dennis Wei, Yunfeng Zhang}

\affiliations{IBM Research}

\usepackage{subcaption}
\usepackage{enumitem}

\usepackage{listings}
\usepackage{lipsum}
\usepackage{courier}
\usepackage{afterpage}
\newcommand{\ignore}[1]{}
\newcommand{\numalgs}{ten }

\newcommand{\myheading}[1]{
\textbf{\emph{#1}}}

\newcommand{\DW}[1]{\textcolor{red}{DW: #1}}

\begin{document}

\maketitle

\begin{abstract}
As artificial intelligence and machine learning algorithms 
become increasingly prevalent in society, 
multiple stakeholders are calling for these algorithms to provide explanations. 
At the same time, these stakeholders, whether they be affected citizens, government regulators, domain experts, or system developers, have different explanation needs. To address these needs, in 2019, we created AI Explainability 360 \cite{aix-jmlr},
an open source software toolkit featuring \numalgs diverse and state-of-the-art explainability methods and two evaluation metrics.  
This paper examines the impact of the toolkit with several case studies, statistics, and community feedback. The different ways in which users have experienced AI Explainability 360 have resulted in multiple types of impact and improvements in multiple metrics,
highlighted by the adoption of the toolkit by the independent LF AI \& Data Foundation.
The paper also describes the flexible design of the toolkit, examples of its use, and the significant
educational material and documentation available to its users.

\end{abstract}

\input{body-new}


\newpage

\fontsize{9.7pt}{10.7pt} \selectfont
\bibliography{aix}

\end{document}

%% file: body-new.tex
\section{Introduction}
\label{sec:intro}

The increasing use of artificial intelligence (AI) systems in high
stakes domains has been coupled with an increase in societal demands
for these systems to provide explanations for their outputs.
This societal demand has already resulted in new regulations
requiring explanations \cite{gdpr-goodman,gdpr-wachter,SelbstP2017,illinois-2019}.
Explanations can allow users to gain insight into the system's
decision-making process, which is a key component in calibrating appropriate trust
and confidence in AI systems~\cite{rsi,Varshney2019}.  

However, many machine learning techniques, which are responsible for
much of the advances in AI, are not easily explainable, even by
experts in the field.  This has led to a growing research community~\cite{KimVW2018},
with a long history, focusing on ``interpretable'' or ``explainable''
machine learning techniques.\footnote{We use the terms explainable and interpretable fairly interchangeably; some scholars make a strong distinction \cite{Rudin2019}.}  However, despite the growing volume of
publications, there remains a gap between what
the research community is producing and how it can be leveraged by society.

One reason for this gap is 
that different people in different settings may require different kinds of explanations~\cite{tomsett-explainability-2018,Hind2019}. 
%
We refer to the people interacting with an AI system as \emph{consumers}, and to their different types as \emph{personas}. For example, a doctor trying to understand an AI
diagnosis of a patient may benefit from seeing known similar cases
with the same diagnosis.
A denied loan applicant will want to
understand the reasons for their rejection and what can be done to
reverse the decision.
A regulator, on the other hand, will want to
understand the behavior of the system as a whole to ensure that it
complies with the law.
A developer may want to understand where
the model is more or less confident as a means of improving its
performance.

As a step toward addressing the gap, in 2019, we released the AI Explainability 360 (AIX360) open source software toolkit for explaining machine learning models and data \citep{aix-jmlr}. The toolkit currently features \numalgs explainability methods (listed in Table~\ref{tab:algs}) and two evaluation metrics from the literature \cite{selfEx,CEM-MAF}. We also introduced a taxonomy to navigate the space of explanation methods, not only the \numalgs in the toolkit but also the broader literature on explainable AI. The taxonomy was intended to be 
usable by consumers with varied backgrounds to choose an appropriate explanation method for their application. 
AIX360 differs from other open source explainability toolkits (see \citet{aix-jmlr} for a list) in two main ways: 1) its support for a broad and diverse spectrum of explainability methods, implemented in a common architecture, and 2) its educational material as discussed below. 



\begin{table*}[t!]
\small
    \centering
    \begin{tabular}{|l|p{10cm}|} \hline
     \emph{BRCG \cite{BDR}}
        & Learns a small, interpretable Boolean rule in disjunctive normal form (DNF) for binary classification.\\ \hline
     \emph{GLRM \cite{GLRM}} 
        & Learns a linear combination of conjunctions for real-valued regression \\
            & through a generalized linear model (GLM) link function (e.g., identity, logit). \\ \hline
     \emph{ProtoDash \cite{proto}} &
        Selects diverse and representative samples that summarize a dataset or explain a test instance. \\
            & Non-negative importance weights are also learned for each of the selected samples.\\ \hline
     \emph{ProfWeight \cite{ProfWeight}} &
        Learns a reweighting of the training set based on a given interpretable model and a high-performing complex neural network.\\
            & Retraining of the interpretable model on this reweighted training set is likely to improve the performance of the interpretable model.\\ \hline
     \emph{TED \cite{TED}} 
        & Learns a predictive model based not only on input-output labels but also on user-provided explanations. \\
            &For an unseen test instance both a label and explanation are returned.\\ \hline
     \emph{CEM \cite{CEM}} 
        & Generates a local explanation in terms of what is minimally sufficient to maintain the original classification, \\
            & and also what should be necessarily absent.\\ \hline
     \emph{CEM-MAF \cite{CEM-MAF}} & 
        For complex images, creates contrastive explanations like CEM above but based on high-level semantically meaningful attributes.\\ \hline
     \emph{DIP-VAE \cite{DIP-VAE}} 
        & Learns high-level independent features from images that possibly have semantic interpretation.\\ \hline

     \emph{LIME \cite{lime}} & 
        Obtains local explanations by fitting a sparse linear model locally. \\
            & The code is integrated from the library maintained by its authors: \url{https://github.com/marcotcr/lime}.\\ \hline
     \emph{SHAP \cite{unifiedPI}} &
        Identifies feature importances based on Shapley value estimation methods. \\
            & The code is integrated from the authors' repository: \url{https://github.com/slundberg/shap}. \\ \hline

    \end{tabular}
    \caption{The AI Explainability 360 toolkit (\texttt{v0.2.1}) includes a diverse collection of explainability algorithms}
    \label{tab:algs}
\end{table*}


The main purpose of the current paper, two years after the release of AIX360, is to look back at the impact that it has had. More specifically

    \noindent \textbf{Impact:} We describe some benefits of the toolkit from our experiences with it, spanning use cases in finance, manufacturing, and IT support, as well as community metrics and feedback. Due to the variety of ways in which others have experienced AIX360 and its algorithms, it has had multiple types of impact: operational, educational, competition, and societal. It has correspondingly brought improvements in multiple metrics: accuracy, semiconductor yield, satisfaction rate, and domain expert time.

\noindent In addition, we discuss aspects of the toolkit and accompanying materials that illustrate its design, ease of use, and documentation. This discussion expands upon the brief summary given in \citet{aix-jmlr}.

     \noindent\textbf{Toolkit Design:} We describe the architecture that enables coverage of a diversity of explainability methods as well as extensions to the toolkit since its initial release. Code listings show how its methods can be easily called by data scientists.

        \noindent \textbf{Educational Material:} We discuss the resources available to make the concepts of explainable AI accessible to nontechnical stakeholders. These include a web demonstration, which highlights how three different personas in a loan application scenario can be best served by different explanation methods. Five Jupyter notebook tutorials show data scientists how to use different methods across several problem domains, including lending, health care, and human capital management. 
       
    \ignore{ 
    \item \emph{Taxonomy Conception:} we propose a simple yet comprehensive taxonomy of AI explainability that considers varied perspectives. This taxonomy is actionable in that it aids users in choosing an approach for a given application and may also reveal gaps in available explainability techniques.
    }


These contributions demonstrate how the design and educational material of the AIX360 toolkit have led to the creation of better 
AI systems.

\ignore{
We have examined the impact of the open source AI Explainability 360 (AIX360) toolkit two years after its initial release. A major motivation for creating the toolkit was that different personas interacting with an AI system have different goals and require different kinds of explanations. This diversity has been borne out in the multiple types of impact that we have discussed, from operational to societal, and in the metrics that have been improved, from accuracy to user satisfaction. We have also discussed how the design of the toolkit supports a range of explanation methods and extensions, given examples of its use, and described the educational material that makes it accessible to practitioners and nonexperts.
}

\ignore{
The rest of the paper is organized as follows. 
Section~\ref{sec:taxonomy} describes our taxonomy and illustrates its simplicity and comprehensiveness. It describes the \numalgs explainability algorithms in the toolkit.
Section \ref{sec:implement} presents 
the toolkit's software architecture, how the \numalgs algorithms are implemented, and the educational material. 
Section~\ref{sec:impact} highlights some of the impact from using the toolkit.
Section \ref{sec:relWork} discusses other related toolkits.
Section \ref{sec:disc} summarizes our contributions and highlights promising directions along which the toolkit can be further enhanced and extended.
}

\section{Initial Impact}
\label{sec:impact}

This section highlights the impact of the AIX360 toolkit in the first two years since its release.  It describes several different forms of impact on real problem domains and the open source community.
This impact has resulted in improvements in multiple metrics: accuracy, semiconductor yield, satisfaction rate, and domain expert time.

The current version of the AIX360 toolkit includes \numalgs explainability algorithms described in Table~\ref{tab:algs} covering different ways of explaining. Explanation methods could be either local or global, where the former refers to explaining an AI model's decision for a single instance, while the latter refers to explaining a model in its entirety. Another dimension of separation is that explanation methods are typically either feature-based or exemplar-based. The former provides key features as an explanation, while the latter provides a list of most relevant instances. Under feature-based, there are also methods termed as contrastive/counterfactual explanations. These provide the most sensitive features which, if slightly changed, can significantly alter the output of the AI model. Another type here are rule-based methods which output semantically meaningful rules and could be considered a subtype of feature-based explanations.

\noindent\myheading{Financial Institution: Educational impact}

\ignore{
A large financial institution formed a dedicated team 
to ensure transparency and trustworthiness of their deployed models.  As with most financial institutions, they employ a separate team to 
validate models built by the model development team. They
}

Large financial institutions typically have dedicated teams to ensure transparency and trustworthiness of their deployed models. These teams validate models built by the model development team. After the release of the AIX360 toolkit, one such financial institution approached us to educate their data science team in a newly founded ``center of excellence''. Based on real use cases seen by this team they created
multiple explainability use cases that varied in modality (tabular, image, and text) and model
types (LSTMs, CNNs, RNNs, boosted trees).  The goal for each use case/model
was to answer the following four questions: 

\ignore{
We recently had a 3-month engagement with a large financial institution that has made a sizeable investment in ensuring transparency and trustworthiness of their deployed deep learning models.
They have formed a dedicated team that validates models built by a different model development team. The bank came to us with around 10 different use cases which spanned all the three modalities (viz. tabular, image and text), and with models that encompassed neural networks with different architectures (viz. LSTMs, CNNs, RNNs) as well as non-differentiable ensembles such as boosted trees. The task for (almost) all of these use cases was to answer the following four questions for a built model: 
}

\begin{description}
\item [Q1:] Generally, what features are the most important for decisions made by the model?
\item [Q2:] What features drove a decision for a certain input?
\item [Q3:] What features could be minimally changed to alter the decision for an input? 
\item [Q4:] Do similar inputs produce the same decision?
\end{description}
These questions provide pragmatic examples of an enterprise's requirement for an explainability technique, which is more concrete than simply ``the model should be able to explain its decision."

\ignore{These questions provide concrete pragmatic examples of what an enterprise might want from an explainability technique, which is a dramatic contrast with a requirement of "the model should be able to explain its decision."}

These questions also represent diversity in the types of explanations needed.
Q1 is a global explainability question. Q2 and Q3 are local feature-based explainability questions, where Q3 is requiring a contrastive explanation. Q4 is a local exemplar-based explainability question. All these questions were answerable through one or more methods available through AIX360: Q1 $\rightarrow$ BRCG, GLRM, ProfWeight; Q2 $\rightarrow$ LIME, SHAP; Q3 $\rightarrow$ CEM, CEM-MAF and Q4 $\rightarrow$ ProtoDash.  In fact, we learned that some of these questions were inspired from our toolkit and the methods it possesses. This demonstrates that not only does the toolkit address many real explainability questions, but it also can help structure thinking about this space in relation to real problems. Thus, its contributions are both technical and  conceptual. 

The result of the engagement was that the data science team was able to successfully test out many of our methods on the different use cases covering the four questions. They came away with newly acquired expertise in this space due in large part to AIX360's existence.

\noindent\myheading{Semiconductor Manufacturing: Operational impact}

\ignore{
 This use case relates to the etching step in the overall process. 
The goal is to predict the quantity of metal etched on each wafer -- which is a collection of chips -- without having to explicitly measure it using high-precision tools. These tools are not only expensive, but also substantially slow down the throughput of a fab. Given a particular specification, we want to predict whether chips lie below this specification by a factor $\alpha > 0$, or above by $\alpha$, or are within spec(ification). We thus have a three class problem. The engineers' goal is not only to predict these classes accurately, but also to obtain insight into ways in which they can improve the process.
}
Semiconductor manufacturing is a multibillion-dollar industry, where producing a modern microprocessor chip is a complex process that takes months.
A semiconductor manufacturer was using a model to estimate the quality of a chip during an etching process, precluding the use of standard tools that are expensive (cost millions of dollars) and time-consuming (can take several days). The engineers' goal was not only to predict quality accurately but also to obtain insight into ways in which they can improve the process. They specifically wanted a decision tree type of model which they were comfortable interpreting. Our goal was thus to build the most accurate ``smallish" decision tree we could.
Using the ProfWeight explanability algorithm~\cite{ProfWeight}, 
we transferred information from an accurate neural network to a decision tree model, elevating its performance by $\approx$ 13\% making it also accurate. We reported the top features: certain pressures, time since last cleaning, and certain acid concentrations. Based on these insights, the engineer started controlling some of them more tightly, improving the total number of within-spec wafers by 1.3\%. 
In this industry a 1\% increase in yield can amount to billions of dollars in savings.

\ignore{
The semiconductor manufacturing engineers with whom we worked had a preference for using decision trees as the model. Given their familiarity with this model, they seemed to understand and trust it more than others. With this preference in mind using ProfWeight \cite{ProfWeight} we were able to deploy not only an improved CART model, but it also produced operationally significant results. We reported the top features based on the improved model to the engineer. These features were certain pressures, time since last cleaning, and certain acid concentrations. Based on these insights, the engineer started controlling the acid concentrations more tightly. This improved the total number of within-spec wafers by 1.3\%. Although this is a small number, it has huge monetary impact, where even a 1\% increase in yield can amount to billions of dollars in savings. The algorithm has been deployed for almost 8 months and they periodically use it to rebuild their decision tree, when they feel the model is no longer capturing the latest trends in the data.\\
}

\noindent\myheading{Information Technology: Operational impact} 

The IT division of a large corporation used a natural language processing (NLP) model to classify
customer complaints into a few hundred categories.  Although this model achieved close to 95\% accuracy, the inability to explain
the misclassifications led to distrust of the system.
The team used the CEM explainability algorithm~\cite{CEM} to compute local explanations.
The experts said that such explanations are highly valuable in providing stakeholders and end-users with confidence in the inner workings of the classifier.  Approximately, 80\% of the explanations of misclassified complaints were deemed reasonable by them compared to 40\% with the previous approach that resembled LIME \cite{lime}. The experts said the algorithm provided much better intuition about why the system made a mistake, showing in most cases that the mistake was acceptable. They felt that this was useful in developing trust in the system. Given the success of our technique for this problem, CEM has been integrated into their multicloud management platform to help accelerate their clients' journey to cloud. Initial parts of the work were also described in a blog post~\cite{ai-it-blog}.

\noindent\myheading{Consumer Finance: Competition impact}

The Fair Isaac Corporation (FICO) is well known for its FICO score, the predominant consumer credit score in the US. FICO organized an Explainable Machine Learning Challenge \cite{FICO2018} around a real-world dataset of home equity line of credit (HELOC) applications. The tasks were to accurately predict whether applicants would satisfactorily repay the HELOC as well as provide local and global explanations.
We used the BRCG algorithm \cite{BDR} 
to produce a remarkably simple, directly interpretable rule set model, consisting of only two rules, each with three conditions. The model's accuracy of $72\%$ was also close to the best achieved by any model of around $74\%$. 

This submission was awarded first place for the highest score in an empirical evaluation. The scores of our submission as well as other submissions were not disclosed.
We do know, however, from a presentation made by FICO that the evaluation involved data scientists with domain knowledge being presented with local and global explanations, without model predictions. They were then asked to predict the model output. Submissions were scored based on a linear combination of the data scientists' predictive accuracy (weight of $70\%$) and time taken ($30\%$). 
Based on this description, 
we conclude that (1) directly interpretable models as provided by AIX360 can offer an appealing combination of accuracy and explainability, especially for regulated industries such as consumer lending, and (2) such models may be preferred by human decision-makers who have to understand and work with them, 
particularly under time constraints.

\noindent\myheading{Regulator: Educational impact} 

Some requirements for AI Explainability come from industry-specific regulation.  Complementary to the previous examples, we were contacted by a large group from a major financial regulator to leverage our expertise in the creation of the toolkit and taxonomy. 
The group wanted to get a deeper understanding of AI explainability
techniques to determine how they should update their explainability regulations for AI models. Financial regulation is trying to ensure credit is extended without taking on unnecessary risk. More accurate models can help achieve this goal, but often they are not used because of existing XAI regulations. 
The group hoped that some of the techniques in the toolkit could be used as a basis for future regulation of financial institutions.

\noindent\myheading{Open Source Community: Societal impact} 

We quantify the impact of AIX360's release in the open source community via three channels: the GitHub repository, PyPI package repository, and the public Slack workspace (\url{aix360.slack.com}). The usage and community-building statistics are given in Table \ref{tab:stats}. 
In addition to these metrics, the three channels provide qualitative evidence of interaction with the community, chiefly problem reports or feature requests on GitHub, and questions about algorithms and event announcements on Slack. 
Another form of engagement comes from public presentations and conference tutorials.  There have been ten presentations, some of which were captured as videos with over 6,000 views.
\begin{table}[htbp]
\vspace{-0.2cm}
\caption{Usage and community statistics as of September 9, 2021 for the AIX360 toolkit, released in August 2019.}
\label{tab:stats}
\centering
  \begin{tabular}{|l|r|} 
  \hline
 Metric & Value \\[0.3ex]\hline
 Forks & 190 \\ %
 Stars & 923 \\  %
 Last 14-day avg. of github views/day & 182 \\  
 Last 14-day avg. of github unique visitors/day & 35.1  \\  
 Last 14-day avg. of unique github clones/day & 2.6 \\  %
 Total PyPI downloads & 26,218 \\
 AIX360 Slack users & 261 \\  %
 Closed pull requests (PRs) & 71\\   %
 Public presentations/tutorials views & 6,849 \\ %
 
 \ignore{
 } 
 \hline
\end{tabular}
\vspace{-0.2cm}
\end{table}

Another measure of impact is the adoption of the toolkit by independent bodies.  One example of this is the LF AI \& Data Foundation's
accepting the toolkit as an incubator project in September, 2020~\cite{lfai-aix-website}.  This open governance organization has over 50 corporate and university members and ``supports open source innovation in artificial intelligence, machine learning, deep learning, and data''~\cite{lfai-website}.

\ignore{
On September 22, 2020, the LF AI Foundation accepted AI Explainability 360 as an incubation project~\cite{lfai-announce,lfai-aix-website}.  
This demonstrates the significant interest shown by the open source community.
}

\ignore{
\textcolor{red}{Massage this text to fit in this section} \DW{I think a paragraph should be written on external contributions, including ongoing ones. I also think it should go before the quotes.}
}

We have encouraged the community to make their own contributions to AIX360. 
A good example of a community contribution is Floid Gilbert's FeatureBinarizerFromTrees, which uses decision trees to binarize features more intelligently for the BRCG and GLRM algorithms. 
Additionally, the authors of the following four papers are currently integrating new algorithms based on counterfactual explanations and influence functions into the toolkit: \citet{lewis, grace, tracinf, sif}. The next version of the  toolkit (\texttt{v0.2.2}) will include these four additional explainability algorithms and is planned to be released by Dec 2021. 

The AIX360 toolkit offers an extensible software architecture and its licensing structure (Apache v2.0) permits free commercial use and distribution of derivative works. This paves the way for enterprises to build new commercial offerings that leverage the toolkit.  Presently, the toolkit is integrated into IBM's Cloud Pak for Data and is the foundational component of a commercial explainability library offered by IBM with additional proprietary algorithms and software support. 


In addition to the above statistics, contributions and extensions, we have also received unsolicated feedback from the broader community via our public slack channel~\cite{aix360-website}.

\begin{quote}
    \textit{``What a fantastic resource (AIX360 is)! Thanks to everyone working on it."}
    
    --- John C. Havens,
    \small{Executive Director of IEEE Global Initiative on Ethics of Autonomous and Intelligent Systems}
\end{quote}

\begin{quote}
    \textit{``I have found aix360 to be most comprehensive."}
    
    --- Arpit Sisodia, \small{Data Scientist with Ericsson}
\end{quote}

\noindent Consistent with their roles, John particularly appreciated our educational material, and Arpit found the toolkit to have the best coverage of a
diverse set of explainability questions.

%
%
%
\begin{figure*}[ht]
  \centering
      \includegraphics[width=0.8\textwidth]{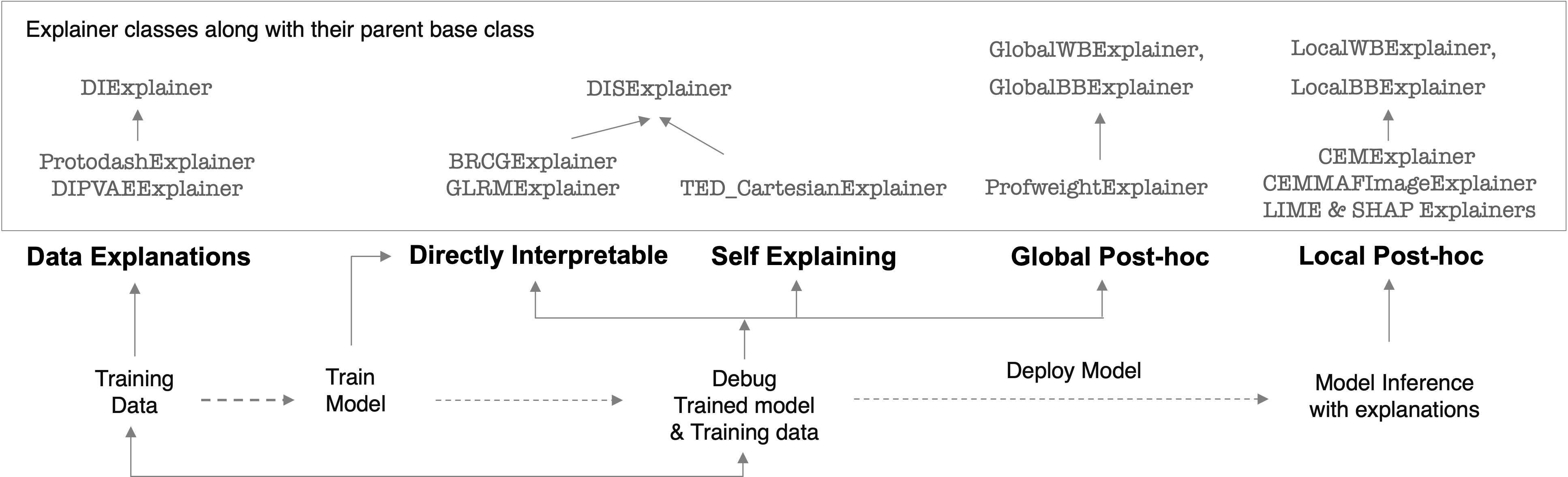}
  \caption{Organization of AIX360 explainer classes according to their use in various steps of the AI modeling pipeline.}
  \label{fig:taxonomy-implementation}
\end{figure*}
\section{AIX360 Design and Usage}
\label{sec:implement}

The AIX360 toolkit offers a unified, flexible, extensible, and easy-to-use programming interface and an associated software architecture to accommodate the diversity of explainability techniques required by various stakeholders. The goal of the architectural design is to be amenable both to data scientists, who may not be experts in explainability, as well as explainability algorithm developers. Toward this end, we make use of a programming interface that is similar to popular Python model development tools (e.g.,~scikit-learn) and construct a hierarchy of Python classes corresponding to explainers for data, models, and predictions. Explainability algorithm developers can inherit from a family of base class explainers to integrate new explainability algorithms. 
The base class explainers are organized according to the AI modeling pipeline shown in Figure~\ref{fig:taxonomy-implementation}, based upon their use in offering explanations at different stages. Below we provide a summary of the various AIX360 classes and illustrate their usage via example in Listing~\ref{lst:dise}.


\begin{itemize}[leftmargin=*]
\item \emph{Data explainers:}  These explainers are implemented using the base class DIExplainer (Directly Interpretable
unsupervised Explainer), which provides abstract methods to implement unsupervised techniques that explain datasets. The AIX360 explainers that inherit from this base class include ProtodashExplainer and DIPVAEExplainer. 

\item \emph{Directly interpretable explainers:} These explainers are implemented using the base class DISExplainer (Directly Interpretable Supervised Explainer), which includes abstract methods to train interpretable models directly from labelled data. The explainers that inherit from this base class and implement its methods include BRCGExplainer and GLRMExplainer. Additionally, the TED\_CartesianExplainer, which trains models using data that is labelled with persona-specific explanations, also inherits from DISExplainer. Listing~\ref{lst:dise}(a) shows an example illustrating the use of BRCGExplainer. 
 
\item \emph{Local post-hoc explainers:} These are further subdivided into black-box and white-box explainers. The black-box explainers are model-agnostic and generally require access only to a model's prediction function.  This class of explainers is implemented via the base class LocalBBExplainer.  Our wrapper implementations of the publicly available LIME and SHAP algorithms inherit from LocalBBExplainer.
The white-box explainers generally require access to a model's internals, such as its loss function, and are implemented via the base class LocalWBExplainer. 
CEMExplainer and CEMMAFImageExplainer both inherit from LocalWBExplainer. 
Listing \ref{lst:dise}(b) shows an example of using the CEMExplainer to obtain pertinent negatives corresponding to MNIST images.

\item \emph{Global post-hoc explainers:} These are subdivided into black-box and white-box explainers as above. The corresponding base classes GlobalWBExplainer and GlobalBBExplainer, include abstract methods that can help train interpretable surrogate models, given a source model along with its data. The ProfweightExplainer, which is an example of a global post-hoc white box explainer, inherits from the base class GlobalWBExplainer.

\item \emph{Dataset and Model API classes:} In addition to explainer classes, AIX360 includes several dataset classes to facilitate loading and processing of commonly used datasets so that users can easily experiment with the implemented algorithms. 

\hspace{0.15cm} To allow users to explain models that have been built using different deep learning frameworks (e.g. TensorFlow, Keras, PyTorch, MXNet) while avoiding the need to implement explainability algorithms multiple times for each framework, AIX360 includes framework-specific classes that expose a common model API needed by explainability algorithm developers. The current version of the toolkt includes model API classes for Keras (based on TensorFlow) and Pytorch models. 



\end{itemize}

\begin{tiny}
\begin{lstlisting}[caption=Example illustrating the use of (a) BRCGExplainer (directly interpretable); (b) CEMExplainer (local post-hoc), label={lst:dise}, language=Python ] 
               (a)      
from aix360.algorithms.rbm import BRCGExplainer, BooleanRuleCG
# Instantiate and train an explainer to compute global rules in conjunctive normal form (CNF)
br = BRCGExplainer(BooleanRuleCG(CNF=True))
br.fit(x_train, y_train)
# print the CNF rules
print (br.explain()['rules'])

               (b)
from aix360.algorithms.contrastive import CEMExplainer
# Instantiate a local post-hoc explainer
explainer = CEMExplainer(mnist_model)
# obtain pertinent negative explanations for a particular image
(pn_image, _, _) = explainer.explain_instance(input_image, arg_mode='PN") 
\end{lstlisting}
\end{tiny}

\section{Educational Material}
AIX360 was developed with the goal of providing accessible resources on explainability to nontechnical stakeholders. Therefore, we include numerous educational materials to both introduce the explainability algorithms provided by AIX360, and to demonstrate how different explainability methods can be applied in real-world scenarios. These educational materials include general guidance for the key concepts of explainability, a taxonomy of algorithms to help a user choose the appropriate one for their use case, a web demo that illustrates the usage of different explainability methods, and multiple tutorials.

\begin{figure}
\vspace{-1.5cm}
  \begin{center}
        (a)\\
     \vspace{2ex}  
      \includegraphics[width=0.45\textwidth]{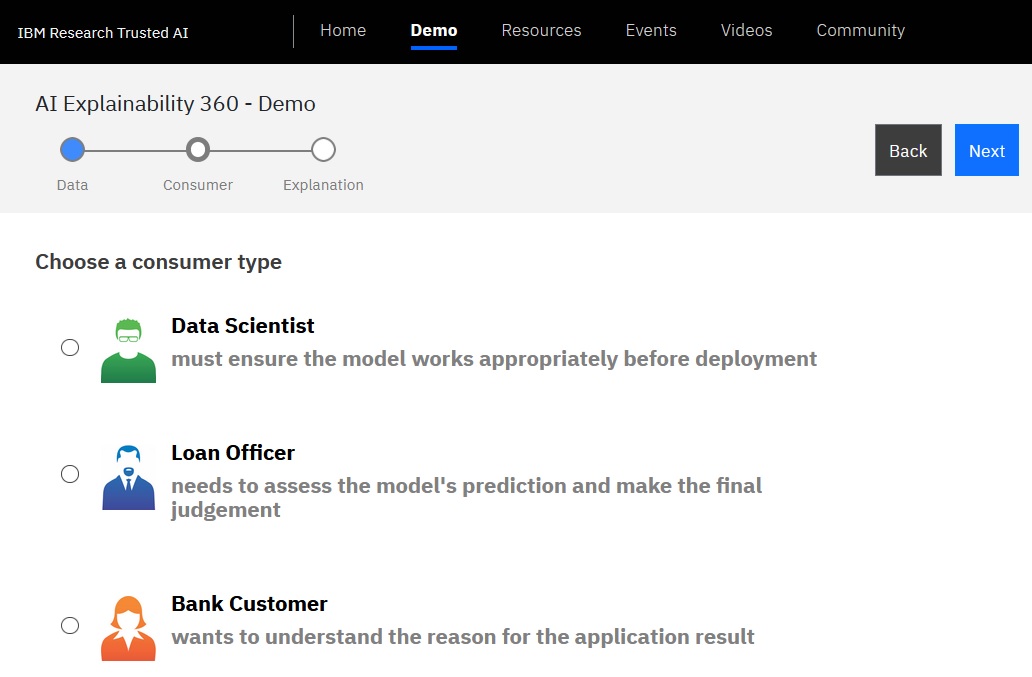}
\vspace{4ex}

        (b)\\
        \vspace{2ex} 
    \includegraphics[width=0.45\textwidth]{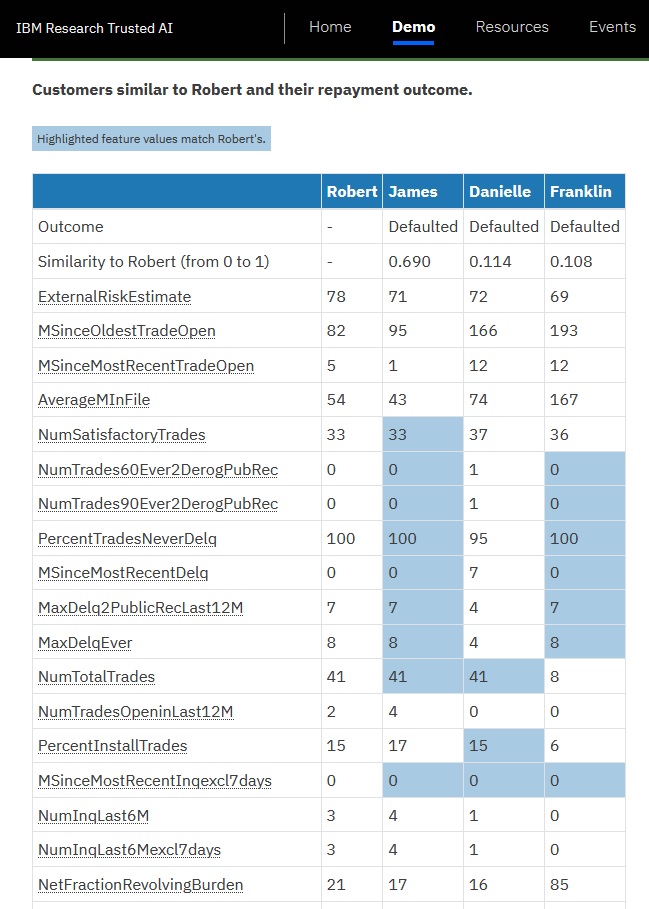}

  \end{center}
  \caption{The web demo illustrates how different types of explanations are appropriate for different personas (image (a)). Image (b) shows
  a subset of the Protodash explainer output to illustrate how similar applicants (i.e. prototypes) in the training data were given the same decision.}
  \label{fig:demo}
  \vspace{-0.5cm}
\end{figure}

A key tenet from our initial work is that ``One Explanation Does Not Fit All'', i.e., different explanation consumers will have different needs, which can be met by different explanation techniques~\cite{aix360-arxiv}.
The web demo~\cite{aix360-website} was created to illustrate this point.  It is based on the  FICO Explainable Machine Learning Challenge dataset \cite{FICO2018}, a real-world scenario where a machine learning system is used to support decisions on loan applications by predicting the repayment risk of the applicants. 
The demo highlights that three groups of people -- data scientists, loan officers, and bank customers -- are involved in the scenario, and their needs are best served by different explainability methods. For example, although the data scientist may demand a global understanding of model behavior through an interpretable model, which can be provided by the GLRM algorithm, a bank customer would ask for justification for their loan application results, which can be generated by the CEM algorithm. We use storytelling and visual illustrations to guide users of AIX360 through these scenarios of different explainability consumers.
Figure~\ref{fig:demo} shows  screenshots from the demo.

The AIX360 toolkit currently includes five tutorials in the form of Jupyter notebooks that show data scientists and other developers how to use different explanation methods across several application domains. The tutorials thus serve as an educational tool and potential gateway to AI explainability for practitioners in these domains. The tutorials cover

    \noindent1. Using three different methods to explain a credit approval model to three types of consumers, based on the FICO Explainable Machine Learning Challenge dataset \cite{FICO2018}.
    
    \noindent2. Creating directly interpretable healthcare cost prediction models for a care-management scenario using Medical Expenditure Panel Survey data.
    
    \noindent3. Explaining dermoscopic image datasets used to train machine learning models by uncovering semantically meaningful features that could help physicians diagnose skin diseases.
    
    \noindent4.  Explaining National Health and Nutrition Examination Survey datasets to support research in epidemiology and health policy by effectively summarizing them.
    
    \noindent5. Explaining predictions of a model that recommends employees for retention actions from a synthesized human resources dataset.

The tutorials not only illustrate the application of different methods but also provide considerable insight into the datasets that are used and, to the extent that these insights generalize, into the respective problem domains. These insights are a natural consequence of using explainable machine learning and could be of independent interest.

\section{Discussion}
\label{sec:disc}

We have examined the impact of the open source AI Explainability 360 (AIX360) toolkit two years after its initial release. A major motivation for creating the toolkit was that different personas interacting with an AI system have different goals and require different kinds of explanations. This diversity has been borne out in the multiple types of impact that we have discussed, from operational to societal, and in the metrics that have been improved, from accuracy to user satisfaction. We have also discussed how the design of the toolkit supports a range of explanation methods and extensions, given examples of its use, and described the educational material that makes it accessible to practitioners and nonexperts.

One lesson we have learned from working with the financial institution in Section~\ref{sec:impact} is the importance of supporting multiple deep learning frameworks, as mentioned in Section~\ref{sec:implement}. The AIX360 toolkit's model API classes helped us explain models trained in different deep learning frameworks without the need to re-implement the explainability algorithms. The model API class currently supports Keras and PyTorch models. We aim to extend it to cover other popular deep learning frameworks in future. 
Another lesson learned is the importance of providing platform-specific installables for air-gapped environments in financial institutions where internet is disabled for security reasons. The library currently supports installation via Pip (package installer for Python), and we plan to provide platform-specific Conda installation packages in future.


\ignore{
\section*{Acknowledgment}

We would like to thank MaryJo Fitzgerald for assistance with the web demo, Karthikeyan Natesan Ramamurthy for assistance with the Medical Expenditure Panel Survey data, and Joanna Brewer for copy editing.
}